\documentclass[sigconf]{acmart}
\usepackage{subfigure}

\AtBeginDocument{%
  }

\setcopyright{acmcopyright}
\copyrightyear{2018}
\acmYear{2018}
\acmDOI{XXXXXXX.XXXXXXX}

\acmConference[Conference acronym 'XX]{Make sure to enter the correct
  conference title from your rights confirmation emai}{June 03--05,
  2018}{Woodstock, NY}
\acmPrice{15.00}
\acmISBN{978-1-4503-XXXX-X/18/06}

\begin{document}

\title{AutoPINN: When AutoML Meets Physics-Informed Neural Networks}

\author{Xinle Wu$^1$, Dalin Zhang$^1$, Miao Zhang$^1$, Chenjuan Guo$^1$, Shuai Zhao$^2$, Yi Zhang$^2$, Huai Wang$^2$, Bin Yang$^1$}
\affiliation{\institution{$^1$Department of Computer Science, Aalborg University}
\institution{$^2$Department of Energy Technology, Aalborg University}
}
\email{{xinlewu, dalinz, miaoz, cguo, byang}@cs.aau.dk, {szh, yiz, hwa}@energy.aau.dk}

\begin{abstract}
Physics-Informed Neural Networks (PINNs) have recently been proposed to solve scientific and engineering problems, where physical laws are introduced into neural networks as prior knowledge. With the embedded physical laws, PINNs enable the estimation of critical  parameters, which are unobservable via physical tools, through observable variables. For example, Power Electronic Converters (PECs) are essential building blocks for the green energy transition. PINNs have been applied to estimate the capacitance, which is unobservable during PEC operations, using current and voltage, which can be observed easily during operations. The estimated capacitance facilitates self-diagnostics of PECs.  

Existing PINNs are often manually designed, which is time-consuming and may lead to suboptimal performance due to a large number of design choices for neural network architectures and hyperparameters. In addition, PINNs are often deployed on different physical devices, e.g., PECs, with limited and varying resources. Therefore, it requires designing different PINN models under different resource constraints, making it an even more challenging task for manual design. To contend with the challenges, we propose Automated Physics-Informed Neural Networks (AutoPINN), a framework that enables the automated design of PINNs by combining AutoML and PINNs. Specifically, we first tailor a search space that allows finding high-accuracy PINNs for PEC internal parameter estimation. We then propose a resource-aware search strategy to explore the search space to find the best PINN model under different resource constraints. We experimentally demonstrate that AutoPINN is able to find more accurate PINN models than human-designed, state-of-the-art PINN models using fewer resources. 
\end{abstract}



\maketitle

\section{Introduction}

A variety of Physics-Informed Neural Networks (PINNs) has recently been proposed to solve Scientific Machine Learning (SciML) problems, achieving better performance than purely data-driven models in many engineering  domains~\cite{karniadakis2021physics,raissi2019physics,mao2020physics,cai2021physics,cai2022physics,misyris2020physics}.
PINN models are typically built by incorporating physical laws as hard or soft constraints into conventional neural networks (NNs), which are then trained normally, e.g., using back-propagation.
For instance, \citet{ji2021stiff} transforms the evolution of the species concentrations in chemical kinetics as the ordinary differential equation, which is then encoded into the loss function to make the network satisfy the governing equations. Since a PINN model embraces certain physical laws, it can be also used to conduct parameter inference from observations, e.g., estimating internal parameters of physical devices that are unobservable via physical instrumentation. A more recent work \cite{zhao2022parameter} builds a PINN to estimate the internal parameters in Power Electronic Converters (PECs), by incorporating the Buck converter physical knowledge into the network training to estimate unknown model coefficients based on the measurable data.

Although the PINN model has been applied to different industrial and physical problems with achieving promising performance, it still has two main limitations. \textbf{First}, existing physics-informed neural networks are manually designed, while the inappropriate network architectures or hyperparameter settings usually lead to the failure of learning relevant physical phenomena for slightly more complex problems \cite{krishnapriyan2021characterizing}. In other words, different neural networks may favour different industrial or physical problems. The range of hyperparameters and possible architectures of neural networks constitute a huge design space so it is unrealistic for even experts to find an optimal PINN model through extensive trials. 
\textbf{Second}, PINN models are often deployed on different physical devices with limited computing capabilities and storage, on e.g., different types of PECs. In addition, different devices may also have quite diverse levels of resources.  
Thus, it is very time-consuming to manually design PINN models that satisfy different hardware constraints.

Recently, Automated Machine Learning (AutoML) has been extensively applied to automatically design machine learning models, and the automatically designed models usually achieve better performance than their manually designed counterparts~\cite{liu2018darts,xu2019pc,DBLP:conf/iclr/ZophL17,pham2018efficient,wu2021autocts,wu2022joint}. This naturally inspires to use AutoML to solve the above two limitations.
In this paper, we propose an Automated Physics-Informed Machine Learning framework (AutoPINN) to design hardware-aware PINN models, where predicting the internal parameters of PECs with our AutoPINN is taken as an exemplary application. Specifically, we first tailor a search space for this task, which contains a wide variety of design choices. Next, we design a reinforcement learning-based search strategy to explore the search space to find the best PINN model. Furthermore, we introduce a hardware-aware reward into the search objective to search for high-accuracy PINN models under given hardware constraints.

To the best of our knowledge, this is the first study integrating AutoML and PINN. The contributions can be summarized as 1) a judiciously designed search space for the task of estimating the internal parameters of PECs with the physics-informed neural network; 2) a reinforcement learning-based search strategy to explore the search space, and a refined search objective to search for the best PINN model under given hardware constraints; 3) extensive experiments on a public dataset, demonstrating that the proposed framework is able to find more accurate and lightweight PINN models than manually designed state-of-the-arts.

\begin{figure*}[!htbp]
\center
\subfigure[PINN]{
\begin{minipage}[c]{0.4\linewidth} 
\centering
\includegraphics[width=\linewidth]{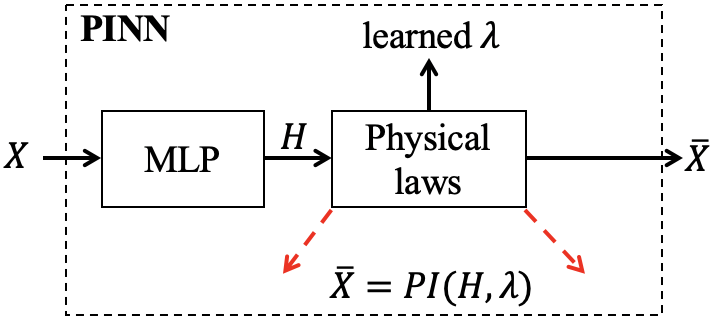}
\label{fig: exist}
\end{minipage}
}
\subfigure[AutoPINN]{
\begin{minipage}[c]{0.53\linewidth} 
\centering
\includegraphics[width=\linewidth]{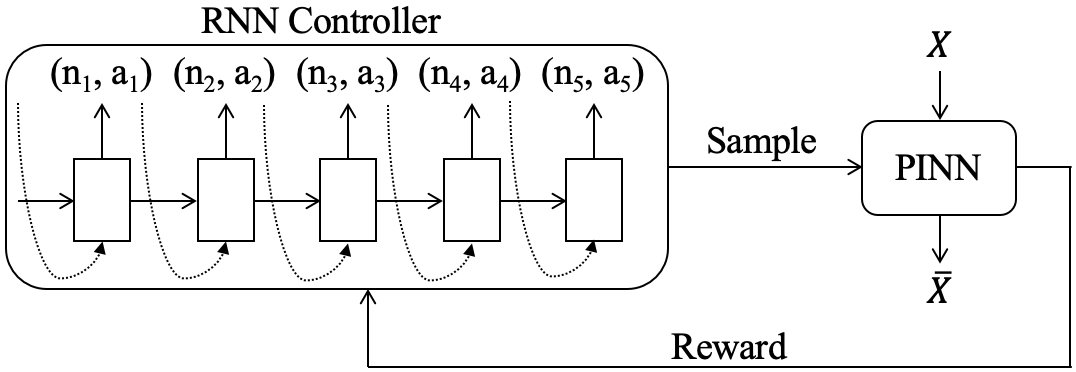}
\label{fig: auto}
\end{minipage}
}
\caption{PINN and AutoPINN. (a) shows the architecture of the existing PINN model. (b) shows the overview of AutoPINN, where $n_i$ denotes the number of units at the $i$-th layer, $a_i$ denotes the activation function at the $i$-th layer.}
\label{fig: overview}
\end{figure*}

\section{Methods}
\label{sec: 2}
\subsection{Physics-informed neural network}
\label{sec: 2.1}

In this paper, we focus on the problem of estimating the internal parameters of Power Electronic Converters (PECs)~\cite{zhao2022parameter}, such as inductance $L$, inductor resistance $RL$, and capacitance $C$, which are important indicators for identifying the aging status of PECs. Accurately estimating these values facilitates early replacement of the PECs that are about to fail.

For this task, a simple data-driven solution is to build a neural network to predict the internal parameters of a PEC with its external observable values (e.g., inductor current and capacitor voltage) as input. This solution demands the measured internal parameters as labels for training the neural network.
However, during PEC operations, we cannot directly measure such interval parameters, because measuring them requires to disassemble the PEC, thus interrupting the normal operations of the PEC. This makes the simple data-driven solution inapplicable.  

PINNs are able to estimate the internal parameters without requiring observing the internal parameters as labels. For example, recent work proposes an autoencoder-paradigm by leveraging a physics-informed neural network to solve this problem~\cite{zhao2022parameter}, which is shown in Figure~\ref{fig: exist}.
It has two components: a pure data-driven component that uses a Multilayer Perceptron (MLP) neural network to encode input observations $X$ (e.g., inductor current and capacitor voltage) into a latent representation $H$, and a physics component that uses physical laws as the decoder $PI$ to decode the latent representation into the evolved observations $\bar{X}$. Once the PINN model converges, the internal parameters $\lambda$ of PECs can be estimated through the learned parameters of the decoder part. More specifically, given the input $X\in R^F$, where $F$ is the feature dimension, the PINN model first encode it to a latent representation $H$ by using an MLP,
\begin{align}
H=MLP(X; \phi),
\label{form: 2.1.1}
\end{align}
where $H\in \mathbb{R}^q$, $q$ is the dimension of the latent representation, 
and $\phi$ is the parameters of the MLP network.
Next, the PINN model can decode the latent representation $H$ in the PEC system to an output $\bar{X}$ by using physical laws,
\begin{align}
\bar{X}=PI(H;\lambda)
\label{form: 2.1.2}
\end{align}
where $PI(\cdot)$ is a physical function with the learned internal parameters $\lambda$ based on the physical law in a PEC system.
The optimization objective is to minimize $|X-\bar{X}|$, formally:
\begin{align}
\lambda^\ast=argmin_{\lambda \in \mathbb{R}^k,\phi}(|X-PI(MLP(X);\lambda)|)
\label{form: 2.1.3}
\end{align}
where $k$ is the number of parameters of the physical function $PI(\cdot)$, and $\lambda^\ast$ is the parameters that lead to the minimal $|X-\bar{X}|$.
The PINN model is trained using backpropagation on the Mean Absolute Error (MAE) loss, and the parameters $\phi$ and $\lambda$ are optimized jointly.

In the task of estimating the internal parameters of PECs, the input $X$ is the vector composed of the inductor current and the capacitor voltage. And the parameters of the physical function $PI(\cdot)$ are also the internal parameters of PECs, that is, $\lambda$.
Therefore, our goal is to obtain the learned $\lambda$ after the PINN model is trained.

\subsection{Automatically designed PINNs}
A drawback of the PINN model in Section~\ref{sec: 2.1} is that we need to manually design the architecture of the neural network.
Due to a large number of choices, it is often challenging in practice to find an appropriate physics-informed neural network through manual network designing.
In addition, real-world industrial applications encounter different hardware constraints under different scenarios, bringing more challenges to designing lightweight PINN models to satisfy different memory and computational constraints.

To tackle this problem, we propose to combine AutoML methods and the PINN model, and aim to automatically design the best PINN model under given hardware constraints.
In this section, we first introduce the search space tailored for estimating the internal parameters of PECs, which contains massive candidate PINN models. And then introduce how we use the reinforcement learning based search strategy to explore the search space. Finally, we introduce the hardware-aware search objective, which supports to search for the best PINN model that satisfies given hardware constraints.

\subsubsection{Search Space}
We adopt MLP to implement the encoder in the PINN model and search for the number of units $n_i$ and the activation function $a_i$ at the $i$-th layer, where $i\leq T$ and $T$ is the maximum number of layers because together they determine the architecture of the MLP and can lead to completely different performance and memory consumption.
The number of units can be chosen from [0, 20, 30, 40, 50, 60], where $0$ means the current layer is discarded. And the activation function can be chosen from [tanh, ReLu]. Therefore, for the $i$-th layer, each element of the possible choice $(n_i, a_i)$ is chosen from the two lists above, respectively.
For example, (20, tanh) is a possible choice for a layer of the MLP, which means the layer consists of 20 units and the activation function of this layer is tanh.

Overall, once the number of units and the activation function at each layer are determined, a PINN model can be built.

\subsubsection{Search Strategy}
To explore the proposed search space, we apply a reinforcement learning based search strategy. 
Figure~\ref{fig: auto} shows the overview of the search strategy. Specifically, we first use a controller to generate the description of PINN models. The controller is implemented by a Recurrent Neural Network (RNN), which works in an autoregressive manner. At the $i$-th ($i\geq 1$) step, the input of the RNN is the output from $(i-1)$-th step, except the first step, whose input is a learned variable. The RNN encodes the input to a hidden state and then inputs it to a softmax classifier to select the candidate PINN models.
At each step, the RNN outputs the number of units and the activation function at each layer.
After the RNN generates all the descriptions, a PINN model is built based on it and trained to obtain the MAE.

Our goal is to learn the parameters of the RNN controller, denoted $\theta$, to ensure that it can generate high-performance PINN models under given hardware constraints. Therefore, we use the reciprocal of the MAE of the generated PINN model as the reward to guide the optimization of the RNN. Specifically, we maximize the following expected reward to learn the parameter $\theta$ of the RNN controller,
\begin{align}
J(\theta)=E_{P(c_{1:T};\theta)}[R]
\label{form: 2.2.2}
\end{align}
where $c_{1:T}$ is the output list of the RNN controller, and $c_{1:T}=\{(n_1, a_1),(n_2, a_2),(n_3, a_3),(n_4, a_4),(n_5, a_5)\}$, and $T=5$ in Figure~\ref{fig: auto}. $P(c_{1:T};\theta)$ is the probability that the controller generate the output $c_{1:T}$. $R=1/MAE$ is computed by training the PINN model built based on $c_{1:T}$, and used as a reward to update the parameter $\theta$ of the RNN controller. The objective encourages the RNN controller to generate PINN models that yield small MAE. 
We follow Zoph et al.~\cite{DBLP:conf/iclr/ZophL17} to use the REINFORCE rule~\cite{williams1992simple} to compute the gradient of $\theta$:
\begin{align}
\bigtriangledown \theta J(\theta)= \sum \limits_{t=1}^{T} \bigtriangledown_{\theta}logP(c_t|c_{(t-1):1};\theta)(1/MAE)
\label{form: 2.2.3}
\end{align}

\subsubsection{Incorporating  Hardware Constraints}
Estimating the internal parameters of PECs is a multi-objective problem in practice, as we usually need to design a PINN model with not only high accuracy but also low memory usage.
Given a storage constraint, our goal is to the best PINN model under the constraint.
To achieve this goal, we modify the search objective in Equation~\ref{form: 2.2.2} so that the search is hardware-aware, formally,
\begin{align}
J(\theta)=E_{P(c_{1:T};\theta)}(1/MAE)([\frac{Param(m)}{P_0}]^w)
\label{form: 2.2.4}
\end{align}
where $Param(\cdot)$ is the function that counts the number of parameters of a PINN model, which is used to approximate the memory usage.
$P_0$ is the given memory constraint, and $w$ is the weighting factor defined as:
$$
w=
\begin{cases}
    \alpha, & if \ Param(m)\leq P_0 \\
    \beta, & otherwise
\end{cases}
\label{form: 2.2.5}
$$
where $\alpha$ and $\beta$ are predefined constants. $m$ represents a PINN model. In this work, we empirically set $\alpha=-0.02$ and $\beta=-0.1$.
The intuition behind this setting is that when the number of parameters of the generated PINN model is less than $P0$, we increase $w$ because in this case, we are more concerned with the accuracy of the searched PINN model. Conversely, when the number of parameters of the generated PINN model is larger than $P0$, we penalize the objective value with a smaller weight to encourage the searched model to satisfy the given hardware constraint.

\begin{table*}[htbp]
    \centering
    \caption{Comparison Results w.r.t. MAE and \# Param, which is the number of NN parameters.}
    \begin{tabular}{c|c|c|cccccccccc}
        \toprule 
        Comparison Models&$Average\: MAE$&\# Param&$L$&$R_L$&$C$&$R_C$&$R_{dson}$&$R_1$&$R_2$&$R_3$&$V_{in}$&$V_F$ \cr
    \midrule  
    \emph{M-PINN}&{5.0}&{12,350}& {0.8}&{13.1}&{1.2}&{4.5}&{27.9}&{0.1}&{0.3}&\textbf{0.1}&{0.1}&{1.9}
    \cr\hline
    \emph{R-PINN-1}&{7.4}&{13,220}& \textbf{0.1}&{12.8}&{5.3}&{4.6}&{29.2}&{0.1}&{0.2}&\textbf{0.1}&{0.3}&{21.9}
    \cr
    \emph{R-PINN-2}&{8.7}&{11,880}& {0.3}&\textbf{12.4}&{8.8}&{4.6}&{29.4}&\textbf{0.0}&\textbf{0.0}&{0.2}&{0.5}&{31.2}
    \cr
    \emph{R-PINN-3}&{8.8}&\textbf{7,620}& {0.4}&\textbf{12.4}&{6.2}&{4.6}&{29.7}&\textbf{0.0}&{0.2}&\textbf{0.1}&{0.5}&{33.4}
    \cr\hline
    \emph{AutoPINN-1}&\textbf{4.7}&{15,480}& {0.9}&{13.0}&{0.8}&\textbf{4.4}&{27.5}&{0.1}&{0.2}&{0.2}&{0.1}&{0.2}
    \cr
    \emph{AutoPINN-2}&{4.9}&{11,220}& {0.9}&{13.4}&\textbf{0.6}&\textbf{4.4}&{28.2}&{0.1}&{0.2}&{0.2}&{0.1}&\textbf{0.1}
    \cr
    \emph{AutoPINN-3}&{5.0}&{9,220}& {0.8}&{12.9}&{1.5}&\textbf{4.4}&\textbf{27.4}&{0.1}&{0.2}&\textbf{0.1}&\textbf{0.0}&{2.9}
    \cr
    \bottomrule
    \end{tabular}
    \label{table}
\end{table*}

\section{Experiments}
\subsection{Experimental Settings}

\noindent\textbf{Dataset. }
To demonstrate the superiority of our proposed AutoPINN, we conduct comparison experiments on a public PEC signal dataset~\cite{zhao2022parameter}.
The data is collected by measuring the transient signals of the {\it inductor current $i$} and {\it output voltage $u$} of a PEC. Only peaks of the inductor current and
output voltage are used. Therefore, the dataset $D$ contains a series of pairs of $i$ and $u$: $D=\{X_k\}$, where $X_k=[i_k, u_k]$. 
The input into the achieved models by AutoPINN and comparison models is $X$, i.e., one pair of $i$ and $u$.
Overall, the dataset contains 360 pairs. For each pair $X_k$, there is a corresponding set of internal parameters $\lambda$, which is measured under laboratory conditions and used as the ground truth for estimating the accuracy of all PINN models. Specifically, $\lambda=\{L,\,R_L,\,C,\,R_C,\,R_{dson},\,R_1,\,R_2,\,R_3,\,V_{in},\,V_F\}$, where $L$ is inductance, $R_L$ is resistance of inductor, $C$ is capacitance, $R_C$ is equivalent series resistance, $R_{dson}$ is onstate resistance, $R_1$, $R_2$ and $R_3$ are three different loads, $V_{in}$ is input voltage, $V_F$ is diode forward voltage, respectively. 
Following~\cite{zhao2022parameter}, we use Mean Absolute Error (MAE) to evaluate the accuracy of PINN models, where lower MAE values indicate higher accuracy.
\medskip

\noindent\textbf{Baselines. }
We compare the AutoPINN with 1) the manually designed PINN model proposed by Zhao et al.~\cite{zhao2022parameter}, denoted as M-PINN, and 2) three randomly selected PINN models from the search space, denoted as R-PINN-1, R-PINN-2, and R-PINN-3, respectively.
We reproduce M-PINN based on the source code released by~\cite{zhao2022parameter}. Note that M-PINN is also embraced in the proposed search space. 
We run AutoPINN three times with different memory constraints to search for three specific PINN models, denoted as AutoPINN-1, AutoPINN-2, and AutoPINN-3, respectively.
\smallskip

\noindent\textbf{Implementation details. }
All experiments are conducted on an Nvidia Quadro RTX 8000 GPU.
We adopt Adam~\cite{kingma2014adam} followed by a full-batch L-BFGS as the optimizer, with the learning rate $\eta$ = 0.001, and exponential decay rates $\beta_1$ = 0.9 and $\beta_2$ = 0.999. The maximum number of MLP layers $T$ is empirically set to 5.
We train all PINN models for 2,000 epochs to convergence and derive the unobservable parameters $\lambda$ from the well-trained PINNs with the help of physical laws.
For AutoPINN-1, AutoPINN-2, and AutoPINN-3, we set their  constraints on parameter numbers to 16,000, 12,000, and 10,000, respectively.

\subsubsection{Experimental Results}
Table~\ref{table} summarizes the MAE performance and model size w.r.t number of parameters of AutoPINN and all baselines.
We use the bold font to highlight the best-performed model in each column.
Key observations are as follows. 

First, the best-performed automatically designed PINN, AutoPINN-1, outperforms the manually designed M-PINN on the average MAE (4.7 vs 5.0). When looking into each parameter individually, we find that AutoPINN-1 performs better or the same as M-PINN on eight out of ten parameters; for the remaining two parameters (i.e., $L$ and $R_3$), AutoPINN-1 is only slightly worse than M-PINN ($L$: 0.9 vs 0.8; $R_3$: 0.2 vs 0.1). In addition, the second-best AutoPINN-2 not only has a lower average MAE than M-PINN (4.9 vs 5.0) but also uses fewer model parameters (11,220 vs 12,350). 
This demonstrates that AutoPINN can find better PINN models than manually designed ones with regard to both accuracy and model size.

Second, when comparing the three automatically generated models, AutoPINN-1 outperforms AutoPINN-2 significantly but also contains extensively more parameters. On the other hand, AutoPINN-3 performs slightly worse than AutoPINN-2 but contains much fewer parameters. Moreover, AutoPINN-3 performs the same as M-PINN regarding the average MAE. This demonstrates that AutoPINN can effectively find well-performed but small PINN models; as the model size increases, the PINN's performance climbs as well.

Third, it is obvious that the randomly sampled PINN models perform extremely poorly. For example, with much more parameters, R-PINN-1 performs far worse than AutoPINN-3. This observation indicates that the proposed search strategy is able to find high-accuracy PINN models from the search space.

\section{Related Work}
\subsection{Physics-informed Machine Learning}

In recent years, there is a large body of work combining neural networks and physical knowledge to solve scientific problems. 
A line of work introduces physical knowledge into the traditional neural network by adding soft constraints to the loss function~\cite{krishnapriyan2021characterizing,chen2020physics,geneva2020modeling,jin2021nsfnets,sahli2020physics}. However, this kind of methods cannot ensure the model strictly follow the physical regulations, which may lead to unreasonable prediction results. On the contrary, several existing works~\cite{greydanus2019hamiltonian,wang2020multi,beucler2019achieving,frerix2020homogeneous,cranmer2020lagrangian} embed specialized physical laws as hard constraints into neural networks, and enforce the model to must satisfy physical constraints. Parameter estimation is another application of PINN \cite{zhao2022parameter,regazzoni2021physics,lakshminarayana2022application,stiasny2021physics}, which can conduct parameter inference from the observations by leveraging linear or nonlinear differential equations. The proposed automated framework is suitable for all PINN models, and we introduce how we combine AutoML and PINN for parameter estimation in this paper.

\subsection{Automated Machine Learning}
There is massive related work on automated machine learning, here we only briefly discuss the most relevant work for this paper. \citet{DBLP:conf/iclr/ZophL17} propose to use an RNN-based controller to produce the description of a neural network and train the controller with reinforcement learning to encourage it to generate high-performance neural networks. Unlike this work, although we also adopt the reinforcement learning based AutoML, we also consider the model size of the produced neural network. \citet{efficientnet} search for small and fast neural networks that can be deployed on mobile devices, and they explicitly incorporate model latency into the search objective. In light of the success of AutoML in various areas, we adopt this thrilling technique to alleviate the costly human efforts in designing PINN models. 
To the best of our knowledge, this paper is the first work to apply AutoML to Physics-informed Machine Learning models.

\section{Conclusion}
We present the AutoPINN to automatically design PINN models for estimating the key but unobservable parameters of power electronic converters. In particular, we first design a search space that contains massive candidate PINN models. Then we apply a reinforcement learning based search strategy to explore the search space to find the best PINN model.
Furthermore, we introduce the model size of the PINN model in the search objective to find the best PINN model under given hardware constraints.
The comparison between the proposed method and baseline methods demonstrates the effectiveness of AutoPINN.

\bibliographystyle{ACM-Reference-Format}
\bibliography{sample-base}

\end{document}